\documentclass{article} %
\usepackage{iclr2027_arxiv,times}

\usepackage{amsmath,amsfonts,bm}

\def\eqref#1{equation~\ref{#1}}

\def\1{\bm{1}}

\DeclareMathAlphabet{\mathsfit}{\encodingdefault}{\sfdefault}{m}{sl}
\SetMathAlphabet{\mathsfit}{bold}{\encodingdefault}{\sfdefault}{bx}{n}

\usepackage{hyperref}
\usepackage{url}
\usepackage{booktabs}
\usepackage{amsfonts}
\usepackage{nicefrac}
\usepackage{microtype}
\usepackage[table]{xcolor}
\usepackage{graphicx}
\usepackage{amsmath}
\usepackage{amssymb}
\usepackage{multirow}
\usepackage{adjustbox}
\usepackage{subcaption}
\usepackage{wrapfig}
\usepackage{siunitx}
\usepackage{algorithm}
\usepackage{algpseudocode}

\title{Dual-QK: Sharp Queries and Flat Keys for Prunable 2-bit KV Caches}

\author{
\makebox[\dimexpr\textwidth-2\tabcolsep\relax][c]{%
\begin{tabular}{c}
Sunjoo Whang$^{1}$ \quad
Jungjun Oh$^{1}$ \quad
Minsung Kim$^{1}$ \quad
Dongho Seo$^{2}$ \\[2pt]
Jisu Shin$^{2}$ \quad
Gregory Kielian$^{3}$ \quad
Hoi-Jun Yoo$^{1}$ \quad
Sangjin Kim$^{2,\dagger}$ \\[4pt]
{\normalfont\small
$^{1}$KAIST \qquad
$^{2}$GIST \qquad
$^{3}$Google Research}
\end{tabular}%
}
}

\newcommand{\dqscore}[2]{%
  \ensuremath{
    \num[round-mode=places,round-precision=1]{#1}
    \pm
    {\scriptstyle\num[round-mode=places,round-precision=1]{#2}}
  }%
}

\newcommand{\dqscorebest}[2]{%
  \ensuremath{
    \mathbf{\num[round-mode=places,round-precision=1]{#1}}
    \pm
    {\scriptstyle\mathbf{\num[round-mode=places,round-precision=1]{#2}}}
  }%
}

\newcommand{\redcell}[1]{\cellcolor{red!8}#1}

\iclrfinalcopy %
\begin{document}

\maketitle

\begingroup
\renewcommand{\thefootnote}{\fnsymbol{footnote}}
\footnotetext[2]{Corresponding author.}
\endgroup

\begin{abstract}
Long inputs and extended generation increase the storage and access
costs of the key-value (KV) cache. Low-bit quantization reduces
storage and memory traffic, while query-channel pruning can further
reduce key-cache reads. Rotation-based quantization redistributes
the energy of key outliers across channels. To maintain computational invariance, the same orthogonal transform
must be applied to queries, preserving query-key dot products.
However, this rotation can disperse query energy, weakening the separation
between a few large components to retain and many small ones to prune. We introduce
Dual-QK, which uses paired non-orthogonal query and key transforms
to address this conflict. Using calibrated query and key statistics,
Dual-QK combines partial key whitening with a query-aligned basis
to balance key scales for INT2 quantization and concentrate query
energy for dynamic channel pruning. Channel-0 protection and
bucket-relative RoPE support low-bit accuracy over long contexts.
Experiments on four models across five generative benchmarks and
long-context retrieval tasks show improved accuracy over OSCAR
on most tasks at 40\% query-channel sparsity. At a 128K context,
Dual-QK provides $6.8\times$ KV-cache compression and an estimated
$8.3\times$ reduction in KV read volume relative to unpruned BF16.
Under the evaluated configurations, our SGLang implementation
achieves up to $3.75\times$ the decoding throughput of unpruned BF16.

\end{abstract}

\section{Introduction}
\label{sec:introduction}

Large language models are expanding from understanding long documents
and code repositories to multistep reasoning and tool use. Long inputs
require substantial context before generation begins, while reasoning
traces and accumulated tool outputs extend the history during generation.
Autoregressive decoding stores this history in the key--value (KV) cache,
and dense attention revisits accumulated keys and values for
every token. As the context grows, storing and repeatedly accessing
this cache places increasing demands on memory capacity and bandwidth.
Efficient long-context inference therefore requires reducing both cache
storage and access costs during decoding.

KV quantization reduces both storage and memory traffic by lowering the number
of bits per element~\citep{kivi,kvquant,oscar}. Dynamic sparse attention
can further reduce reads by selecting tokens or channels for each query
while retaining the cached history~\citep{quest,sparq}. Each query--key (QK)
dot product is a sum of channel contributions, so omitting channels with
small contributions reduces the key components needed to compute
attention scores. Combining query-channel sparsity with low-precision
keys allows fewer channels to be read at fewer bits, preserving the
storage benefits of quantization while further reducing key accesses.
Realizing both benefits requires a representation that supports accurate
key quantization and effective channel selection.

\begin{figure}[t]
    \centering
    \includegraphics[width=\linewidth]{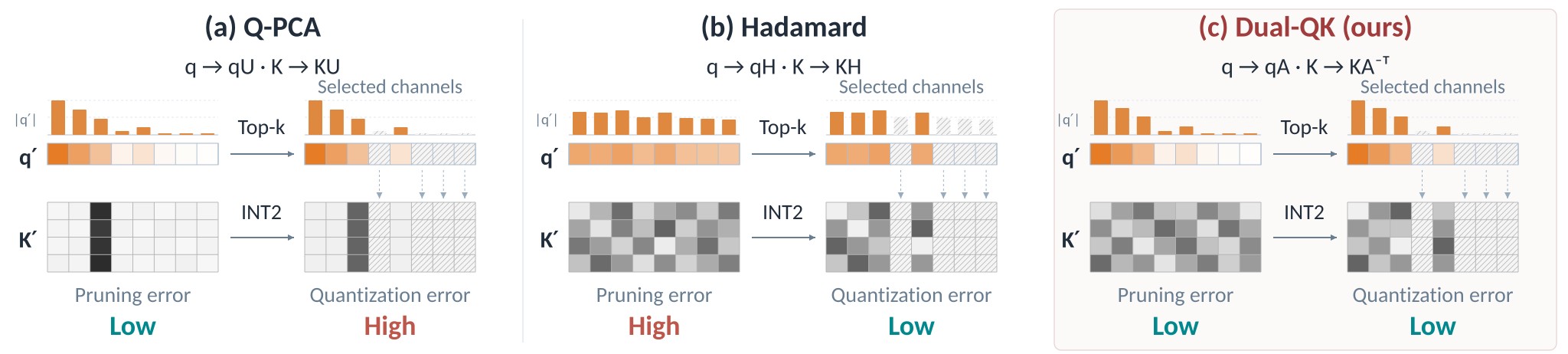}
    \caption{
    Conceptual comparison at the same bit width and retained-channel
    budget. (a) Q-PCA concentrates query energy, whereas (b) Hadamard rotation
    mitigates key outliers. (c) Dual-QK targets both properties using
    reciprocal Q/K transforms. Hatched channels are omitted from the
    dot product.
    }
    \label{fig:qk_tradeoff}
\end{figure}

However, key quantization and query-channel pruning favor different channel
distributions, as illustrated in Figure~\ref{fig:qk_tradeoff} (a) and (b). For key
quantization, outliers widen the shared quantization range and reduce
the resolution available to other values, favoring a flatter distribution.
Query-channel pruning instead benefits from a sharp distribution that
separates a few important channels from many small ones.
Rotation-based quantization redistributes the energy of key outliers
across channels~\citep{quarot,spinquant,oscar}. To maintain computational
invariance, the same orthogonal transform is applied to queries,
preserving query--key dot products before quantization. However,
this shared transform can also disperse query energy, weakening
the concentration exploited by pruning.
Conversely, query principal component analysis (Q-PCA) uses the
eigenvectors of query second moments as a shared rotation basis
for queries and keys. This concentrates query energy in a few
channels but does not explicitly control key outliers that affect
quantization error.
These contrasting requirements motivate a joint design of query and key
representations.

We propose \textbf{Dual-QK}, which uses paired non-orthogonal query
and key transforms to address this conflict. Pairing an invertible
query transform with its inverse transpose on keys preserves
query--key dot products before quantization and pruning. This extends
the shared orthogonal rotations used in prior work, allowing the
query and key distributions to be shaped differently. Dual-QK uses
this freedom to balance key scales while concentrating query energy
in a few channels. These transforms support 2-bit key quantization
and dynamic channel selection based on current query magnitudes.

Specifically, Dual-QK combines partial key whitening with a query-aligned
basis derived from calibrated query and key statistics to balance key
scales and concentrate query energy. Large query magnitudes make attention scores sensitive to quantization error in the leading key channel (channel-0). We therefore retain this channel in BF16 and exclude it from range estimation to narrow the quantization range of the remaining INT2 channels.
Bucket-relative RoPE represents each stored key relative to its bucket
anchor, keeping stored-key positional offsets within the calibration
window. The corresponding query adjustment preserves attention scores
before quantization and pruning.

Our contributions are as follows:
\begin{itemize}
    \item We analyze the conflicting distributional requirements of key
    quantization and query-channel pruning and introduce Dual-QK, a paired
    transformation that constructs flat keys and sharp queries while
    preserving attention scores before compression.

    \item We exploit these distributions to combine dynamic query-channel
    selection with INT2 key quantization, using leading-channel protection
    and bucket-relative RoPE to maintain accuracy across context positions.

    \item We evaluate Dual-QK on four models for generative accuracy,
    long-context retrieval, and decoding throughput.
    At 40\% query-channel sparsity, Dual-QK improves accuracy over OSCAR
    on most tasks. At a 128K context, it provides $6.8\times$
    KV-cache compression and an estimated $8.3\times$ reduction in KV
    read volume relative to unpruned BF16.
\end{itemize}

\section{Related Work}
\label{sec:background}

\subsection{KV Cache Quantization}
\label{sec:background_quantization}

KV cache quantization reduces storage and memory
traffic by storing keys and values at lower precision.
Quantization accuracy depends on how values share a quantization range
and how outliers are handled. Mixed precision, fine-grained grouping,
and channel rescaling address these issues in LLM
quantization~\citep{llmint8,atom,qserve}. For KV caches,
KIVI~\citep{kivi} quantizes keys per channel and values per token.
KVQuant~\citep{kvquant} combines per-channel key quantization before
rotary position embedding (RoPE) with non-uniform quantization and
separate outlier storage. Other approaches improve low-bit KV compression through channel
reordering and group-wise clipping~\citep{skvq}, or estimate
query--key inner products using sign-quantized random
projections~\citep{qjl}.

Rotation-based methods have achieved state-of-the-art results in 4-bit
LLM quantization by redistributing outliers through orthogonal
transforms~\citep{quarot,spinquant}, with outlier-aware variants targeting
KV caches~\citep{rotatekv}. TurboQuant~\citep{turboquant} combines random
rotations with non-uniform scalar quantization for KV-cache compression.
OSCAR~\citep{oscar} combines calibrated spectral bases with Hadamard
mixing for INT2 KV quantization. Serving-oriented methods integrate KV
quantization with efficient attention kernels~\citep{qserve,sawint4}.

\subsection{Sparse Attention}
\label{sec:background_sparsity}

KV cache eviction reduces both storage and read traffic but may
discard information needed by future queries. Query-aware sparse
attention methods retain the full cache
and select relevant entries for each query, reducing read traffic
while keeping unselected information available for later
queries~\citep{sparq,quest}. This flexibility can better preserve
accuracy as token importance changes across queries~\citep{quest}.

\noindent\textbf{Token-level sparsity.}
Token-based methods retain or evict cached tokens using attention
statistics and positional heuristics~\citep{h2o,streamingllm,snapkv,pyramidkv}.
Query-aware access selects KV pages using key statistics~\citep{quest}
or predicts upcoming attention to prefetch relevant entries from host
memory~\citep{infinigen}. LServe~\citep{lserve} combines KV quantization
with static streaming attention and dynamic page selection.

\noindent\textbf{Channel-level sparsity.}
Channel-based methods select dimensions of query--key dot products.
ThinK~\citep{think} uses prompt query--key statistics to remove
low-importance channels from older keys and reuses the mask during
decoding. SparQ Attention, Double Sparsity, and Loki use selected
channels or low-dimensional key representations to approximate token
importance before attending to retrieved
tokens~\citep{sparq,doublesparsity,loki}. Related approaches reduce
dimensionality through low-rank
projections~\citep{eigenattention,palu,matryoshkakv},
which can be combined with
quantization~\citep{palu,matryoshkakv}.
Dual-QK jointly designs Q/K coordinates for low-bit key
quantization while keeping all transformed key channels
available for dynamic selection based on current query magnitudes.

\section{Method}
\label{sec:method}

\subsection{Dual Transforms}
\label{sec:dual_transform}

We use row-vector notation and denote queries and keys after RoPE by
$Q$ and $K$, with head dimension $d$. Transforming keys changes their
dot products unless queries receive a compensating transform.
For any invertible matrix $A$, choosing $B=A^{-\mathsf T}$ gives
\begin{equation}
    (QA)(KB)^{\mathsf T}=QK^{\mathsf T}.
    \label{eq:qk_computational_invariance}
\end{equation}
When $A$ is orthogonal, $B=A$, recovering a shared rotation for queries and keys. Dual-QK allows
the two transforms to differ, so key scales can be adjusted without
forcing the same adjustment onto queries. This identity preserves the
original scores before quantization and pruning.

For each layer and KV head, we estimate the uncentered second moments
$C_Q=\mathbb E[q^{\mathsf T}q]$ and
$C_K=\mathbb E[k^{\mathsf T}k]$ from calibration data. Query statistics
are pooled over heads sharing the KV head. We first reduce scale
differences across key directions using partial whitening. With
$G=C_K+\epsilon I$ and $\epsilon>0$, keys are multiplied by
$G^{-\alpha/2}$ and queries by its inverse, $G^{\alpha/2}$. The parameter
$\alpha\in[0,1]$ controls whitening strength; regularization limits
amplification along directions with little calibration energy.

\begin{figure}[t]
    \centering
    \includegraphics[width=\linewidth]{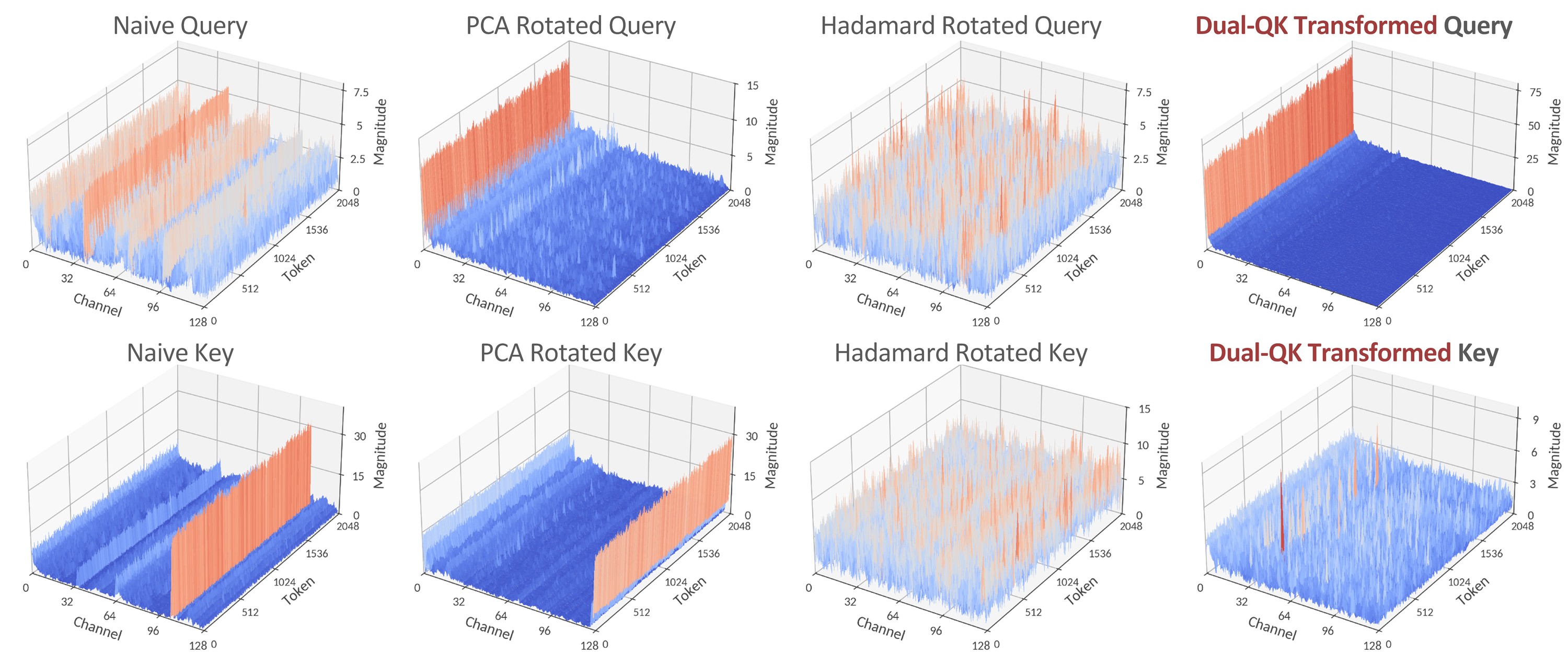}
    \caption{
    Query (top) and key (bottom) magnitudes from Llama-3.1-8B-Instruct
    (layer 29, KV group 6, query head 26) in the original, Q-PCA,
    Hadamard, and Dual-QK coordinates. Q-PCA concentrates queries but
    leaves key imbalance. Hadamard spreads both, whereas Dual-QK
    concentrates queries while reducing key imbalance. Magnitude axes
    are scaled independently across panels.
    }
    \label{fig:qk_distributions}
\end{figure}

Whitening alone does not arrange query energy for pruning. We therefore
rotate the compensated queries into their principal directions:
\begin{equation}
    G^{\alpha/2}C_QG^{\alpha/2}
    =U\Lambda U^{\mathsf T},
    \label{eq:compensated_query_moment}
\end{equation}
where the diagonal entries of $\Lambda$ are ordered from largest to
smallest. Applying the same orthogonal matrix $U$ to queries and keys
gives the base transforms
\begin{equation}
    A_0=G^{\alpha/2}U,
    \qquad
    B_0=G^{-\alpha/2}U.
    \label{eq:dual_pair}
\end{equation}
The transformed query moment is
$A_0^{\mathsf T}C_QA_0=\Lambda$: leading channels capture the dominant
calibrated query energy. Meanwhile, partial whitening compresses the
scale differences in the regularized key moment. At $\alpha=0$, the
pair reduces to Q-PCA; at $\alpha=1$, the base transform whitens the
regularized key moment. Increasing whitening strength also changes the
query spectrum through the reciprocal scaling. Partial whitening
controls how much the two distributions are reshaped, balancing key
scale reduction with query-energy concentration.
Figure~\ref{fig:qk_distributions} illustrates the resulting distributions.
These moment properties describe average channel energy on the
calibration distribution; dynamic selection accounts for variation
across individual queries.

We choose $\alpha$ once per model and reuse the transforms across
benchmarks and context lengths. The matrix decompositions are performed
offline. During inference, each layer and KV head uses a fixed transform
pair, while the channel mask adapts to the current queries. The moment
identities and calibration procedure are given in
Appendices~\ref{app:dual_geometry} and~\ref{app:calibration}.

\subsection{Dynamic Query-Channel Pruning}
\label{sec:channel_pruning}

The calibrated ordering identifies channels that are important on
average, but their importance changes with each query. We therefore
select channels dynamically. Query magnitude is a useful proxy when
key channels have comparable scales: otherwise, a small query component
can still multiply a large key component. Key flattening makes this
proxy more appropriate, although it does not make query energy an exact
measure of attention-score error.

In grouped-query attention (GQA)~\citep{gqa}, $g$ query heads share one
KV head. Let $q'_h=q_hA_0$ be the base transformed query of head $h$.
For each channel $j$, we compute the L2 norm across the group:
\begin{equation}
    e_j=\sqrt{\sum_{h=1}^{g}(q'_{h,j})^2}.
    \label{eq:dual_channel_selection}
\end{equation}
We retain the $m$ channels with the largest $e_j$ and apply the same mask
to all query heads in the group. This maximizes the retained query energy
for the chosen channel budget. Multi-head attention (MHA) is the case
$g=1$, where selection reduces to query magnitude.

Attention dot products use the retained channels and keep the original
normalization by $\sqrt d$. All key channels remain stored, allowing a
channel skipped by one query to be used by the next. With the channel-major layout in Section~\ref{sec:kv_quantization}, the shared mask lets the kernel load only selected key channels. Quantization reduces bits per element, while pruning reduces the key elements read. Values and metadata are still read.

\subsection{INT2 KV Quantization}
\label{sec:kv_quantization}

In the base coordinates, a key error $\delta k'_j$ on a retained
channel contributes $q'_{h,j}\delta k'_j/\sqrt d$ to head $h$'s
attention logit. The leading base query coordinate has the largest calibrated expected
energy, making errors in its corresponding key coordinate especially
consequential for attention scores. We therefore retain key channel~0
in BF16 and exclude it from the quantization range of the remaining
channels. Those channels use one asymmetric INT2 grid per token and
KV head. Removing channel~0 from range estimation also prevents its
magnitude from reducing the resolution available to the other channels.
Precision protection does not force channel~0 into every dot product;
it remains subject to the dynamic mask.

Second moments do not fully describe the ranges used by a low-bit
quantizer. We further refine channel scales with a positive diagonal
matrix $D$, using $A=A_0D$ and $B=B_0D^{-1}$. Inspired by reciprocal
channel scaling~\citep{qserve}, we calibrate $D$ to reduce the
attention-score error caused by key quantization and fix its channel-0
entry to one. The two scale factors cancel in each channel. Thus,
rescaling preserves both the full dot product and the contribution of
any fixed channel set in the base dual coordinates before quantization.
The refinement is folded into the transforms before inference. Channel
selection still uses the base queries $q_hA_0$; retained dot products
use the final scaled queries and keys.

For values, we follow OSCAR's calibrated rotation, clipping, and INT2
quantization recipe~\citep{oscar}, reversing the value rotation after
aggregation. We retain the first 64 and most recent 256 tokens in BF16
for both keys and values. These tokens use the same query-channel mask.
Quantization details appear in appendix.

Within each 128-token page, INT2 keys are stored channel-major
for each KV head, with one channel occupying a contiguous
32-byte row. The decode kernel issues asynchronous copies
only for rows selected by the shared query mask.
Values are stored token-major, with separate buffers for
quantization metadata and BF16-protected entries.

\subsection{Bucket-Relative RoPE}
\label{sec:bucket_rope}

The dual basis is calibrated over a finite positional window. At longer
positions, RoPE changes the channel statistics presented to this fixed
basis. We address this mismatch by expressing stored keys relative to
local bucket anchors. This changes their positional frame without
recalibrating the transforms.

We divide positions into buckets of width $w$, with bucket $b$ starting
at $a_b=bw$. Let $T_b$ undo the RoPE rotation at this anchor. A key $k_n$
in the bucket and a current query $q$ are represented as
\begin{equation}
    z_n=k_nT_bB,
    \qquad
    \widetilde q_b=qT_bA.
    \label{eq:bucket_coordinates}
\end{equation}
Because $T_b$ is orthogonal and the dual transforms cancel, their
uncompressed dot product is unchanged:
\begin{equation}
    \widetilde q_b z_n^{\mathsf T}=qk_n^{\mathsf T}.
    \label{eq:bucket_score_preservation}
\end{equation}
The stored key's residual position is $n-a_b\in[0,w)$. We use $w=2048$,
matching the calibration window, and reuse the same dual pair for every
bucket. This bounds the positional offsets of stored keys; the query's
relative position can still exceed the window. The anchor adjustment precedes the dual transform so that the
calibrated basis acts on bucket-relative coordinates. General
channel transforms do not commute with RoPE, so reversing this
order changes the representation used for quantization and
channel selection.

During decoding, each new key is transformed in its bucket frame and
stored. For each bucket, the current query heads receive the corresponding
anchor adjustment, and Equation~\ref{eq:dual_channel_selection} is
applied to their base coordinates $q_hT_bA_0$. Thus, channel masks can
differ across buckets. The selected channels produce attention scores
using the final scaled coordinates. Scores from all buckets and precision
tiers share one global softmax, followed by value aggregation and the
inverse value rotation.

\section{Experiments}
\label{sec:experiments}

\subsection{Experimental Settings}
\label{sec:experimental_settings}

\noindent \textbf{Models and benchmarks.}
We evaluate Dual-QK on Llama-3.1-8B-Instruct~\citep{llama3},
Qwen3-4B-Thinking-2507, Qwen3-8B~\citep{qwen3}, and
Ministral-3-14B-Reasoning-2512~\citep{ministral3}.
Generative benchmarks cover knowledge reasoning with GPQA-Diamond,
code generation with HumanEval and LiveCodeBench v6, and mathematical
reasoning with MATH-500 and AIME25~\citep{gpqa,humaneval,livecodebench,math500, aime_2025}.
We report pass@1 for code generation and accuracy for reasoning tasks.
Long-context retrieval is evaluated on the needle-in-a-haystack (NIAH)
tasks of RULER~\citep{ruler}, with context lengths from 4K to 128K.
All methods use the same prompts, sampling parameters, generation
limits, and scoring procedure within each model--benchmark pair.
Repetition counts are specified in the corresponding table and
figure captions.

\noindent \textbf{Baselines and comparison protocol.}
We compare Dual-QK with KIVI~\citep{kivi} and two rotation-based
quantization methods, TurboQuant~\citep{turboquant} and OSCAR~\citep{oscar}.
All model weights remain in BF16. For the main accuracy comparison,
we apply the same query-channel pruning policy at 40\% sparsity
to all quantized methods and use unpruned BF16 as the reference.
Following OSCAR, Dual-QK retains the KV entries of the first 64
and most recent 256 tokens in BF16.
At 128K context, we report format-based estimates of KV storage
and read volume as effective bits per KV element (BPE).
Capacity BPE describes the stored cache, while bandwidth BPE
estimates the KV data read per decoding step with channel selection.
Both account for quantization metadata and high-precision entries,
including the protected sink and recent tokens.

\noindent \textbf{Calibration and implementation.}
Dual-QK is calibrated on 128 sequences of 2,048 tokens from the
WikiText-2 training set~\citep{wikitext2} and we select $\alpha$ using separate 8K and 32K inputs. We collect post-RoPE query
and key second moments for each layer and KV head, pooling query
statistics across heads that share the same KV head.
The transforms are constructed offline without retraining model
weights and reused across benchmarks and context lengths.
We implement Dual-QK in SGLang~\citep{sglang}.
We evaluate accuracy on NVIDIA H200 GPUs and decoding throughput
on a single NVIDIA RTX PRO 6000 Blackwell GPU with 96 GB of memory.

\subsection{Main Results}

\begin{table}[t]
\caption{Generative accuracy (\%) with 40\% query-channel sparsity
for quantized methods. BF16 is unpruned. Scores report mean $\pm$
standard deviation over five seeds (three for Ministral-3).
Capacity and bandwidth are effective bits per KV element at
128K context. \textbf{Bold} marks the highest mean among quantized methods
within each model.}
\label{tab:generative}
\footnotesize
\centering
\setlength{\tabcolsep}{1.5pt}
\renewcommand{\arraystretch}{0.9}
\resizebox{\linewidth}{!}{%
\begin{tabular}{cl|cc|ccccc}
\toprule
\multirow{2}{*}{\textbf{~~Model~~}}
& \multirow{2}{*}{\textbf{Method}}
& \multicolumn{2}{c|}{\textbf{KV bits per element}}
& \multicolumn{5}{c}{\textbf{Benchmark score (\%)}} \\
\cmidrule{3-9}
& & \textbf{Capacity} & \textbf{Bandwidth}
& \textbf{GPQA-Diamond} & \textbf{HumanEval}
& \textbf{LCB v6} & \textbf{MATH-500} & \textbf{AIME25} \\
\midrule

\multirow{5}{*}{%
    \fontsize{8}{8}\selectfont
    \shortstack{\textbf{Llama-3.1-8B}\\\textbf{-Instruct}}%
}
& BF16 & 16 & 16
& \dqscore{26.2}{1.5}
& \dqscore{66.7}{2.5}
& \dqscore{16.5}{1.4}
& \dqscore{49.4}{1.5}
& \dqscore{1.3}{1.8} \\
& TurboQuant & 3.13 & 2.53
& \dqscore{17.7}{2.4}
& \dqscore{24.3}{1.4}
& \dqscore{3.8}{0.5}
& \dqscore{13.8}{1.1}
& \dqscore{0.7}{1.5} \\
& KIVI & 3.01 & 2.41
& \dqscore{19.2}{2.7}
& \dqscore{55.2}{4.3}
& \dqscore{9.8}{2.6}
& \dqscore{28.9}{1.4}
& \dqscore{0.7}{1.5} \\
& OSCAR & 2.28 & 1.88
& \dqscore{16.4}{3.0}
& \dqscore{28.2}{2.2}
& \dqscore{3.2}{1.0}
& \dqscore{16.4}{0.4}
& \dqscorebest{1.3}{1.8} \\
\rowcolor{red!8}
\cellcolor{white}
& \textbf{Dual-QK (ours)} & 2.34 & 1.93
& \dqscorebest{23.2}{1.9}
& \dqscorebest{64.3}{2.9}
& \dqscorebest{14.2}{1.3}
& \dqscorebest{43.0}{2.0}
& \dqscore{0.7}{1.5} \\
\midrule

\multirow{5}{*}{%
    \fontsize{8}{8}\selectfont
    \shortstack{\textbf{Qwen3-4B}\\\textbf{-Thinking-2507}}%
}
& BF16 & 16 & 16
& \dqscore{65.6}{1.9}
& \dqscore{96.2}{0.7}
& \dqscore{54.0}{1.6}
& \dqscore{98.0}{0.3}
& \dqscore{74.7}{3.8} \\
& TurboQuant & 3.13 & 2.53
& \dqscore{3.5}{1.6}
& \dqscore{0.0}{0.0}
& \dqscore{0.0}{0.0}
& \dqscore{0.2}{0.2}
& \dqscore{0.0}{0.0} \\
& KIVI & 3.01 & 2.41
& \dqscore{54.3}{2.8}
& \dqscore{88.4}{1.8}
& \dqscore{31.6}{2.1}
& \dqscore{92.5}{1.0}
& \dqscore{42.0}{6.9} \\
& OSCAR & 2.28 & 1.88
& \dqscore{23.1}{2.1}
& \dqscore{33.9}{1.7}
& \dqscore{2.1}{1.1}
& \dqscore{51.8}{1.6}
& \dqscore{6.0}{1.5} \\
\rowcolor{red!8}
\cellcolor{white}
& \textbf{Dual-QK (ours)} & 2.34 & 1.93
& \dqscorebest{60.8}{2.2}
& \dqscorebest{95.0}{0.8}
& \dqscorebest{38.8}{2.3}
& \dqscorebest{94.3}{0.4}
& \dqscorebest{48.0}{5.6} \\
\midrule

\multirow{5}{*}{%
    \fontsize{8}{8}\selectfont
    \shortstack{\textbf{Qwen3-8B}}%
}
& BF16 & 16 & 16
& \dqscore{56.1}{2.2}
& \dqscore{90.1}{1.3}
& \dqscore{50.2}{2.9}
& \dqscore{96.4}{0.3}
& \dqscore{65.3}{3.0} \\
& TurboQuant & 3.13 & 2.53
& \dqscore{19.8}{2.9}
& \dqscore{16.0}{1.2}
& \dqscore{2.4}{0.8}
& \dqscore{31.7}{1.7}
& \dqscore{0.0}{0.0} \\
& KIVI & 3.01 & 2.41
& \dqscore{49.0}{3.2}
& \dqscore{87.3}{1.8}
& \dqscore{32.4}{2.9}
& \dqscore{92.0}{1.0}
& \dqscore{47.3}{4.9} \\
& OSCAR & 2.28 & 1.88
& \dqscore{38.1}{3.2}
& \dqscore{66.1}{4.3}
& \dqscore{14.8}{2.4}
& \dqscore{80.3}{0.8}
& \dqscore{13.3}{4.1} \\
\rowcolor{red!8}
\cellcolor{white}
& \textbf{Dual-QK (ours)} & 2.34 & 1.93
& \dqscorebest{55.7}{1.9}
& \dqscorebest{90.6}{2.0}
& \dqscorebest{36.2}{1.2}
& \dqscorebest{95.0}{0.3}
& \dqscorebest{53.3}{2.4} \\
\midrule

\multirow{5}{*}{%
    \fontsize{8}{8}\selectfont
    \shortstack{\textbf{Ministral-3-14B}\\\textbf{-Reasoning-2512}}%
}
& BF16 & 16 & 16
& \dqscore{53.7}{1.1}
& \dqscore{96.3}{1.2}
& \dqscore{55.7}{0.8}
& \dqscore{95.4}{0.7}
& \dqscore{43.3}{5.8} \\
& TurboQuant & 3.13 & 2.53
& \dqscore{44.8}{4.9}
& \dqscore{93.3}{2.7}
& \dqscore{40.7}{2.2}
& \dqscore{85.4}{0.9}
& \dqscore{23.3}{0.0} \\
& KIVI & 3.01 & 2.41
& \dqscore{45.1}{0.8}
& \dqscore{92.3}{1.8}
& \dqscore{41.5}{1.2}
& \dqscore{86.4}{0.2}
& \dqscore{27.8}{6.9} \\
& OSCAR & 2.28 & 1.88
& \dqscore{37.4}{2.5}
& \dqscore{91.5}{0.6}
& \dqscore{37.2}{1.2}
& \dqscore{85.5}{0.8}
& \dqscore{25.6}{1.9} \\
\rowcolor{red!8}
\cellcolor{white}
& \textbf{Dual-QK (ours)} & 2.34 & 1.93
& \dqscorebest{50.2}{3.2}
& \dqscorebest{96.3}{0.6}
& \dqscorebest{47.6}{3.6}
& \dqscorebest{92.7}{0.4}
& \dqscorebest{37.8}{1.9} \\
\bottomrule
\end{tabular}%
}
\end{table}

\noindent \textbf{Generative benchmarks.}
Table~\ref{tab:generative} shows that Dual-QK maintains higher
generative accuracy under 40\% query-channel pruning at a storage
cost comparable to OSCAR. Dual-QK uses 2.34 capacity BPE, compared
with 2.28 for OSCAR, while improving accuracy on most knowledge,
coding, and mathematical reasoning tasks. On Qwen3-4B-Thinking-2507,
HumanEval pass@1 reaches 95.0\%, compared with 33.9\% for OSCAR
and 96.2\% for unpruned BF16. Dual-QK also achieves higher mean accuracy than KIVI on most tasks while using 2.34 rather than 3.01 capacity BPE.
Among the evaluated quantized methods, Dual-QK generally remains
closest to the BF16 reference.

\begin{table}[t]
\centering
\caption{RULER NIAH accuracy (\%) across context lengths with 40\% query-channel sparsity for quantized methods. BF16 is the unpruned reference. Entries report mean $\pm$ sample standard deviation across three NIAH data draws, each containing 25 examples per subtask across eight subtasks. Capacity and bandwidth are effective bits per KV element at a 128K context. \textbf{Bold} denotes the highest mean among quantized methods.}
\label{tab:ruler_niah}
\footnotesize
\setlength{\tabcolsep}{1.5pt}
\renewcommand{\arraystretch}{0.9}
\resizebox{\linewidth}{!}{%
\begin{tabular}{cl|cc|cccccc}
\toprule
\multirow{2}{*}{\textbf{~~Model~~}}
& \multirow{2}{*}{\textbf{Method}}
& \multicolumn{2}{c|}{\textbf{KV bits per element}}
& \multicolumn{6}{c}{\textbf{Context length}} \\
\cmidrule{3-10}
& & \textbf{Capacity} & \textbf{Bandwidth}
& \textbf{4K} & \textbf{8K} & \textbf{16K}
& \textbf{32K} & \textbf{64K} & \textbf{128K} \\
\midrule

\multirow{3}{*}{%
    \fontsize{8}{8}\selectfont
    \shortstack{\textbf{Llama-3.1-}\\\textbf{8B-Instruct}}%
}
& BF16 & 16 & 16
& \dqscore{99.958}{0.072}
& \dqscore{99.875}{0.000}
& \dqscore{99.958}{0.072}
& \dqscore{99.958}{0.072}
& \dqscore{99.292}{0.260}
& \dqscore{94.833}{0.402} \\
& OSCAR & 2.28 & 1.88
& \dqscore{65.042}{1.018}
& \dqscore{53.583}{1.563}
& \dqscore{51.250}{0.433}
& \dqscore{42.875}{1.682}
& \dqscore{39.542}{1.751}
& \dqscore{25.208}{0.591} \\
& \redcell{\textbf{Dual-QK (ours)}}
& \redcell{2.34} & \redcell{1.93}
& \redcell{\dqscorebest{92.375}{0.375}}
& \redcell{\dqscorebest{92.500}{0.760}}
& \redcell{\dqscorebest{89.000}{0.217}}
& \redcell{\dqscorebest{83.667}{0.688}}
& \redcell{\dqscorebest{78.917}{0.641}}
& \redcell{\dqscorebest{58.875}{1.474}} \\
\midrule

\multirow{3}{*}{%
    \fontsize{8}{8}\selectfont
    \shortstack{\textbf{Qwen3-4B}\\\textbf{-Thinking}\\\textbf{-2507}}%
}
& BF16 & 16 & 16
& \dqscore{100.000}{0.000}
& \dqscore{98.958}{0.144}
& \dqscore{98.333}{0.402}
& \dqscore{96.833}{0.402}
& \dqscore{96.708}{0.564}
& \dqscore{94.000}{0.125} \\
& OSCAR & 2.28 & 1.88
& \dqscore{78.375}{0.820}
& \dqscore{72.000}{0.375}
& \dqscore{61.167}{0.439}
& \dqscore{54.958}{0.832}
& \dqscore{32.708}{1.583}
& \dqscore{2.000}{0.000} \\
& \redcell{\textbf{Dual-QK (ours)}}
& \redcell{2.34} & \redcell{1.93}
& \redcell{\dqscorebest{94.000}{0.250}}
& \redcell{\dqscorebest{92.292}{0.144}}
& \redcell{\dqscorebest{88.833}{0.191}}
& \redcell{\dqscorebest{84.750}{0.433}}
& \redcell{\dqscorebest{81.458}{0.722}}
& \redcell{\dqscorebest{64.375}{0.217}} \\
\midrule

\multirow{3}{*}{%
    \fontsize{8}{8}\selectfont
    \shortstack{\textbf{Qwen3-8B}}%
}
& BF16 & 16 & 16
& \dqscore{99.917}{0.072}
& \dqscore{99.750}{0.000}
& \dqscore{99.958}{0.072}
& \dqscore{98.958}{0.072}
& \dqscore{86.667}{0.191}
& \dqscore{76.542}{1.325} \\
& OSCAR & 2.28 & 1.88
& \dqscore{87.125}{0.217}
& \dqscore{80.042}{0.878}
& \dqscore{73.958}{0.946}
& \dqscore{55.125}{1.566}
& \dqscore{43.042}{0.887}
& \dqscore{10.750}{0.500} \\
& \redcell{\textbf{Dual-QK (ours)}}
& \redcell{2.34} & \redcell{1.93}
& \redcell{\dqscorebest{95.625}{0.331}}
& \redcell{\dqscorebest{95.375}{0.125}}
& \redcell{\dqscorebest{91.875}{0.125}}
& \redcell{\dqscorebest{83.500}{0.661}}
& \redcell{\dqscorebest{68.375}{0.696}}
& \redcell{\dqscorebest{57.000}{1.409}} \\
\midrule

\multirow{3}{*}{%
    \fontsize{8}{8}\selectfont
    \shortstack{\textbf{Ministral-3-}\\\textbf{14B-Reasoning}\\\textbf{-2512}}%
}
& BF16 & 16 & 16
& \dqscore{99.708}{0.144}
& \dqscore{97.750}{0.125}
& \dqscore{98.625}{0.217}
& \dqscore{91.292}{0.629}
& \dqscore{92.125}{0.650}
& \dqscore{82.917}{0.887} \\
& OSCAR & 2.28 & 1.88
& \dqscore{93.333}{0.473}
& \dqscore{89.750}{0.451}
& \dqscorebest{85.750}{0.331}
& \dqscore{76.292}{0.955}
& \dqscore{73.125}{0.331}
& \dqscore{65.625}{1.192} \\
& \redcell{\textbf{Dual-QK (ours)}}
& \redcell{2.34} & \redcell{1.93}
& \redcell{\dqscorebest{95.167}{0.289}}
& \redcell{\dqscorebest{90.417}{0.473}}
& \redcell{\dqscore{83.167}{0.289}}
& \redcell{\dqscorebest{78.708}{0.439}}
& \redcell{\dqscorebest{77.833}{1.041}}
& \redcell{\dqscorebest{71.125}{1.192}} \\
\bottomrule
\end{tabular}%
}
\end{table}

\noindent \textbf{Long-context retrieval.}
Table~\ref{tab:ruler_niah} reports NIAH retrieval accuracy from 4K
to 128K context lengths. Dual-QK improves over OSCAR at most lengths
across all four models, with particularly large gains on Llama and
the two Qwen models. At 64K, Llama-3.1-8B-Instruct and
Qwen3-4B-Thinking-2507 reach 78.9\% and 81.5\%, compared with
39.5\% and 32.7\% for OSCAR. The advantage extends to 128K,
where Dual-QK retains 57.0$\sim$64.4\% accuracy across Llama and
the two Qwen models, compared with 2.0--25.2\% for OSCAR.
Ministral shows smaller improvements at most lengths, although
OSCAR remains higher at 16K. These gains use the same 40\%
query-channel sparsity and comparable KV storage and read budgets.
The selected transforms remain fixed throughout the context-length sweep.

\subsection{Ablation Studies}
\label{sec:ablations}

\begin{figure}[t]
    \centering
    \includegraphics[width=\linewidth]{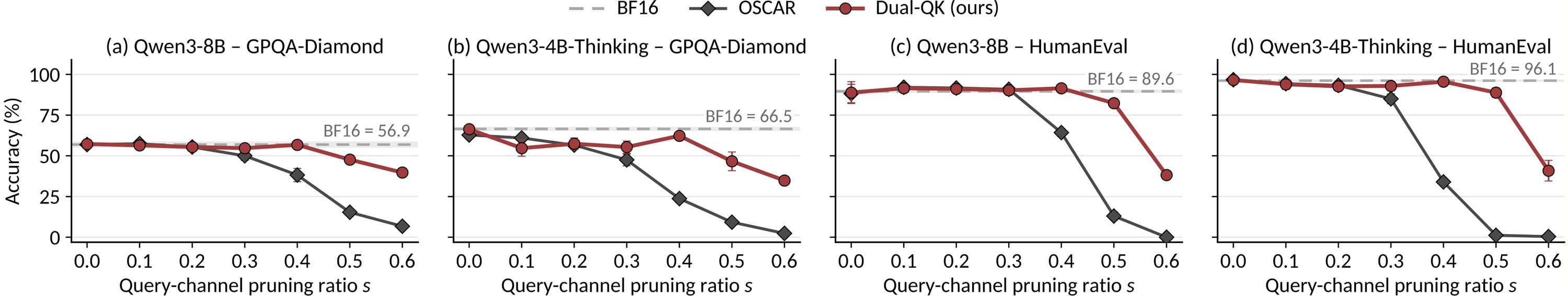}
    \caption{Accuracy on GPQA-Diamond and HumanEval as query-channel
    pruning increases for Qwen3-8B and Qwen3-4B-Thinking-2507.
    Dashed lines indicate the unpruned BF16 references.
    Solid curves show OSCAR and Dual-QK with quantized KV caches,
    including at zero pruning. Error bars indicate standard deviation over three seeds.}
    \label{fig:pruning_sweep}
\end{figure}

\noindent\textbf{Sensitivity to channel pruning.}
Figure~\ref{fig:pruning_sweep} varies the query-channel pruning ratio
from 0\% to 60\% on GPQA-Diamond and HumanEval for Qwen3-8B and
Qwen3-4B-Thinking-2507. Dual-QK and OSCAR perform similarly at
low sparsity, but the accuracy gap widens as more channels are
removed. At 40\%, Dual-QK retains most of its accuracy without
pruning on both tasks, while OSCAR declines substantially.
Dual-QK remains ahead on both tasks and models at 50\%.
At 60\%, its HumanEval accuracy also drops sharply.
These results show that accuracy under quantization alone
does not guarantee robustness to channel pruning.

\begin{table}[t]
\centering
\caption{Component ablation of Dual-QK on Qwen3-8B with INT2 KV
quantization and 40\% query-channel sparsity.
BF16 is unpruned.
Entries report mean $\pm$ standard deviation.
BF16 and Full Dual-QK reuse the five-seed results from
Table~\ref{tab:generative}.
The remaining configurations use three seeds.}
\label{tab:component_ablation}
\begingroup
\footnotesize
\setlength{\tabcolsep}{3pt}
\renewcommand{\arraystretch}{0.9}
\resizebox{0.8\linewidth}{!}{%
\begin{tabular}{lccccc}
\toprule
\textbf{Configuration} & \textbf{GPQA-D} & \textbf{HumanEval} & \textbf{LCB v6} & \textbf{MATH-500} & \textbf{AIME25} \\
\midrule
BF16 reference
& \dqscore{56.1}{2.2}
& \dqscore{90.1}{1.3}
& \dqscore{50.2}{2.9}
& \dqscore{96.4}{0.3}
& \dqscore{65.3}{3.0} \\
\midrule
\rowcolor{red!8}
\textbf{Full Dual-QK} ($\alpha=0.5$)
& \dqscore{55.7}{1.9}
& \dqscore{90.6}{2.0}
& \dqscore{36.2}{1.2}
& \dqscore{95.0}{0.3}
& \dqscore{53.3}{2.4} \\
\quad w/o channel-0 protection
& \dqscore{46.5}{4.5}
& \dqscore{71.5}{1.9}
& \dqscore{16.5}{1.2}
& \dqscore{87.3}{1.0}
& \dqscore{24.4}{1.9} \\
\quad w/o bucket-relative RoPE
& \dqscore{33.2}{0.3}
& \dqscore{51.4}{2.5}
& \dqscore{8.4}{0.8}
& \dqscore{51.1}{1.1}
& \dqscore{2.2}{1.9} \\
\quad w/o diagonal refinement ($D=I$)
& \dqscore{52.5}{4.8}
& \dqscore{87.6}{2.8}
& \dqscore{35.4}{1.9}
& \dqscore{89.5}{5.4}
& \dqscore{51.1}{9.6} \\
Q-PCA base ($\alpha=0$)
& \dqscore{45.3}{3.8}
& \dqscore{75.2}{3.9}
& \dqscore{15.3}{1.3}
& \dqscore{78.7}{11.6}
& \dqscore{17.8}{1.9} \\
Hadamard base ($H$)
& \dqscore{7.9}{1.1}
& \dqscore{4.5}{2.1}
& \dqscore{0.3}{0.4}
& \dqscore{9.8}{2.6}
& \dqscore{0.0}{0.0} \\
\bottomrule
\end{tabular}%
}
\endgroup
\end{table}

\noindent\textbf{Component ablation.}
Table~\ref{tab:component_ablation} compares component removals and
alternative Q/K bases on Qwen3-8B at 40\% query-channel sparsity.
Q-PCA ($\alpha=0$) lowers LiveCodeBench accuracy from 36.2\% to
15.3\% and AIME25 from 53.3\% to 17.8\%. These results indicate
that concentrating query energy alone is insufficient for joint
quantization and pruning. The Hadamard base yields larger
accuracy losses. Removing channel-0 protection or bucket-relative
RoPE lowers HumanEval accuracy from 90.6\% to 71.5\% and 51.4\%,
respectively. Removing diagonal refinement lowers the mean
accuracy on all five benchmarks, although variability across
seeds is substantial on MATH-500 and AIME25.

\begin{table}[t]
\centering
\caption{Generative accuracy (\%) \textbf{without query-channel pruning}.
Benchmark entries report mean $\pm$ standard deviation over five seeds.
BF16 is the unquantized reference.
\textbf{Bold} denotes the highest score among quantized methods within each model.}
\label{tab:zero_sparsity}
\begingroup
\footnotesize
\setlength{\tabcolsep}{2pt}
\renewcommand{\arraystretch}{0.9}
\resizebox{0.8\linewidth}{!}{%
\begin{tabular}{clcccccc}
\toprule
\textbf{Model}
& \textbf{Method}
& \textbf{GPQA-D}
& \textbf{HumanEval}
& \textbf{LCB v6}
& \textbf{MATH-500}
& \textbf{AIME25}
& \textbf{Mean} \\
\midrule

\multirow{4}{*}{%
    \shortstack{\textbf{Llama-3.1-}\\\textbf{8B-Instruct}}%
}
& BF16
& \dqscore{26.2}{1.5}
& \dqscore{66.7}{2.5}
& \dqscore{16.5}{1.4}
& \dqscore{49.4}{1.5}
& \dqscore{1.3}{1.8}
& 32.0 \\
\cmidrule{2-8}
& TurboQuant
& \dqscorebest{26.8}{3.1}
& \dqscore{59.3}{2.9}
& \dqscore{12.5}{1.3}
& \dqscore{39.9}{1.1}
& \dqscore{0.0}{0.0}
& 27.7 \\
& OSCAR
& \dqscore{24.3}{1.7}
& \dqscore{65.6}{3.0}
& \dqscore{15.3}{0.9}
& \dqscore{50.8}{0.5}
& \dqscorebest{2.0}{1.8}
& 31.6 \\
\rowcolor{red!8}
\cellcolor{white}
& \textbf{Dual-QK (ours)}
& \dqscore{26.4}{1.9}
& \dqscorebest{66.0}{1.7}
& \dqscorebest{16.5}{0.7}
& \dqscorebest{50.9}{0.6}
& \dqscore{1.3}{3.0}
& \textbf{32.2} \\
\midrule

\multirow{4}{*}{\textbf{Qwen3-8B}}
& BF16
& \dqscore{56.1}{2.2}
& \dqscore{90.1}{1.3}
& \dqscore{50.2}{2.9}
& \dqscore{96.4}{0.3}
& \dqscore{65.3}{3.0}
& 71.6 \\
\cmidrule{2-8}
& TurboQuant
& \dqscore{47.1}{2.2}
& \dqscore{75.7}{1.6}
& \dqscore{24.7}{1.7}
& \dqscore{92.4}{0.6}
& \dqscore{50.7}{2.8}
& 58.1 \\
& OSCAR
& \dqscorebest{56.6}{1.2}
& \dqscorebest{92.2}{1.1}
& \dqscorebest{46.1}{1.6}
& \dqscorebest{95.8}{0.5}
& \dqscorebest{61.3}{1.8}
& \textbf{70.4} \\
\rowcolor{red!8}
\cellcolor{white}
& \textbf{Dual-QK (ours)}
& \dqscore{55.9}{2.2}
& \dqscore{92.0}{0.9}
& \dqscore{43.2}{2.4}
& \dqscore{95.2}{0.4}
& \dqscorebest{61.3}{3.0}
& 69.5 \\
\bottomrule
\end{tabular}%
}
\endgroup
\end{table}

\noindent\textbf{Accuracy without channel pruning.}
Table~\ref{tab:zero_sparsity} evaluates quantization accuracy without
pruning. Dual-QK achieves a mean accuracy across the five
benchmarks of 32.2\% on Llama-3.1-8B-Instruct and 69.5\% on
Qwen3-8B, compared with 31.6\% and 70.4\% for OSCAR, respectively.
The means differ by less than one percentage point on each model.
OSCAR achieves a higher mean on Qwen3-8B, with the largest gap
on LiveCodeBench.
Together with Figure~\ref{fig:pruning_sweep}, these results show
that Dual-QK maintains competitive accuracy without pruning
while supporting higher channel pruning ratios.

\begin{table}[t]
\centering
\caption{Calibration-domain ablation of Dual-QK on Qwen3-4B-Thinking-2507. Results are reported as mean $\pm$ standard deviation over three seeds. Sparsity denotes query-channel sparsity. BF16 is the unpruned reference. \textbf{Bold} highlights WikiText, our default calibration dataset, and its results.}
\label{tab:calibration_domain}
\begingroup
\footnotesize
\setlength{\tabcolsep}{3pt}
\renewcommand{\arraystretch}{0.9}
\resizebox{0.8\linewidth}{!}{%
\begin{tabular}{lccccccc}
\toprule
\textbf{Method} & \textbf{Sparsity} & \textbf{Calibration} & \textbf{GPQA-D} & \textbf{HumanEval} & \textbf{LCB v6} & \textbf{MATH-500} & \textbf{AIME25} \\
\midrule
BF16 & \textemdash & \textemdash
& \dqscore{63.0}{2.4}
& \dqscore{95.5}{0.9}
& \dqscore{51.7}{0.4}
& \dqscore{97.5}{0.3}
& \dqscore{72.2}{3.8} \\
\midrule
\shortstack{\textbf{Dual-QK}}
& \multirow{3}{*}{0\%} & \textbf{WikiText}
& \dqscorebest{62.3}{1.6}
& \dqscorebest{94.9}{0.4}
& \dqscorebest{44.5}{0.9}
& \dqscorebest{96.1}{0.2}
& \dqscorebest{47.8}{1.9} \\
(\textbf{ours}) & & MMLU
& \dqscore{61.3}{2.8}
& \dqscore{94.7}{0.7}
& \dqscore{45.3}{2.9}
& \dqscore{95.7}{0.4}
& \dqscore{57.8}{1.9} \\
& & GPQA-D
& \dqscore{62.3}{1.8}
& \dqscore{94.9}{0.9}
& \dqscore{45.0}{1.5}
& \dqscore{96.7}{0.8}
& \dqscore{60.0}{5.8} \\
\cmidrule{2-8}
& \multirow{3}{*}{40\%} & \textbf{WikiText}
& \dqscorebest{58.1}{1.3}
& \dqscorebest{92.9}{1.8}
& \dqscorebest{38.9}{2.3}
& \dqscorebest{93.9}{0.8}
& \dqscorebest{51.1}{1.9} \\
& & MMLU
& \dqscore{56.6}{1.5}
& \dqscore{94.7}{0.4}
& \dqscore{39.4}{1.6}
& \dqscore{93.7}{1.0}
& \dqscore{50.0}{0.0} \\
& & GPQA-D
& \dqscore{57.7}{1.1}
& \dqscore{94.9}{2.1}
& \dqscore{41.0}{1.6}
& \dqscore{94.2}{0.4}
& \dqscore{55.6}{1.9} \\
\bottomrule
\end{tabular}%
}
\endgroup
\end{table}

\noindent\textbf{Calibration-domain sensitivity.}
Table~\ref{tab:calibration_domain} compares WikiText-2, MMLU, and
GPQA-Diamond calibration on Qwen3-4B-Thinking-2507 at 0\% and
40\% query-channel sparsity. At 40\%, the spread across calibration
datasets is 0.5 percentage points on MATH-500, 2.0 on HumanEval,
and 2.1 on LiveCodeBench. GPQA-Diamond calibration does not yield
a higher GPQA-Diamond mean than WikiText-2 at either sparsity.
AIME25 shows a larger spread of 12.2 points without pruning and
5.6 points at 40\%. WikiText-2 calibration is competitive on most
tasks, while AIME25 is more sensitive to the calibration source.

\subsection{Throughput}
\label{sec:throughput}

\begin{figure}[t]
    \centering
    \includegraphics[width=\linewidth]{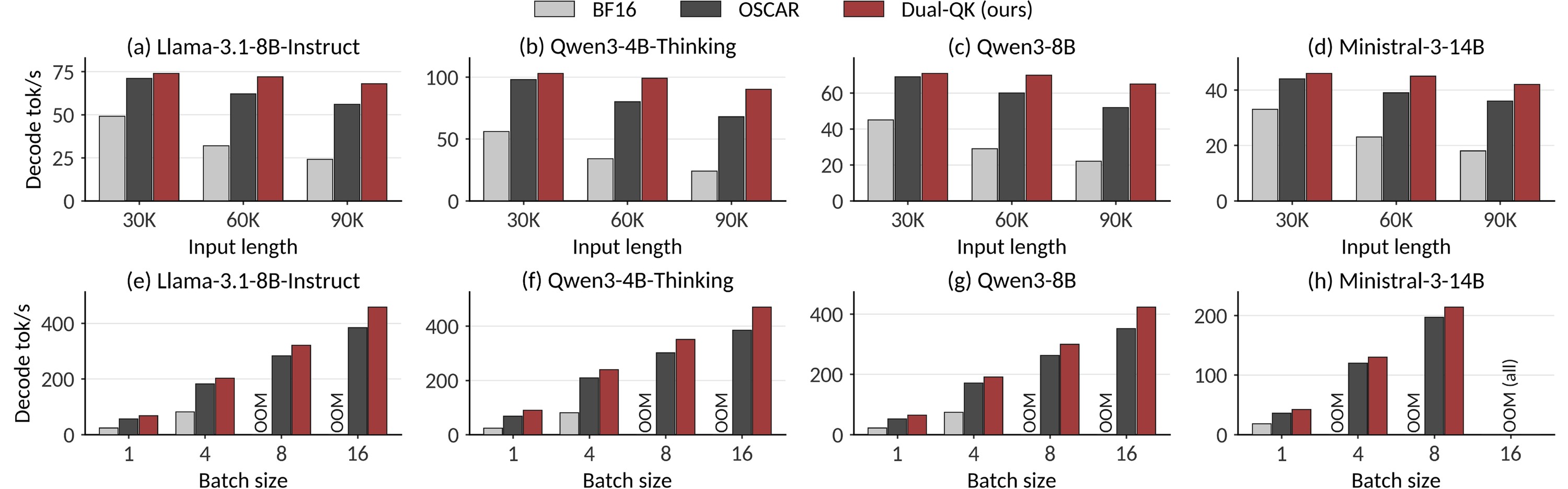}
    \caption{Decoding throughput excluding prefill.
    Top: throughput versus input length @ batch size = 1.
    Bottom: throughput versus batch size @ input length = 90K.
    BF16/OSCAR are unpruned; Dual-QK uses 40\% query-channel sparsity.
    OOM indicates insufficient GPU memory.
}
    \label{fig:decode_throughput}
\end{figure}

\noindent \textbf{Decoding Speedup.} Figure~\ref{fig:decode_throughput} compares BF16, OSCAR, and
Dual-QK across all four models. At batch size 1, Dual-QK achieves
higher throughput than both baselines at every tested input length
under the evaluated configurations. Its relative advantage over
OSCAR grows with input length as accessing the KV cache becomes
more costly. On Qwen3-4B-Thinking-2507 at 90K, Dual-QK reaches
90 tokens/s compared with 24 tokens/s for BF16, giving a
$3.75\times$ speedup. These measurements include transformations
and cache operations during decoding. 

\noindent \textbf{Effect of Batch Size.} The batch sweep shows how cache compression increases serving
capacity within the same GPU memory budget. At 90K, BF16 runs
out of memory at batch sizes 8 and 16 on Llama and both Qwen
models, while OSCAR and Dual-QK support batch size 16.
On Ministral, BF16 runs out of memory starting at the tested
batch size of 4, while both quantized methods support batch
size 8. All three methods run out of memory at batch size 16.
Across the feasible batch sizes shown, Dual-QK maintains higher
aggregate throughput than OSCAR.

\section{Discussion}
\noindent\textbf{Practical use.}
KV-cache quantization enables larger batches on a single GPU.
As weight accesses are amortized across more requests, KV-cache
reads can become the bottleneck. Channel pruning can then
provide additional decoding speedup.

\noindent \textbf{Limitations.} The additional speedup from pruning is limited at small batch
sizes, where selection overhead can offset the savings from
skipped key reads. Its benefit grows when KV-cache access accounts
for more of the decode step, as in larger batches and longer
contexts. Accuracy gaps to BF16 remain at long contexts.
Bucket-relative RoPE bounds stored-key offsets, but query offsets
relative to older buckets can exceed the 2K calibration window.
We leave calibration over broader positional ranges to future work.

\section{Conclusion}
\label{sec:conclusion}

We introduce Dual-QK, a paired query--key transformation for combining low-bit KV quantization with dynamic query-channel pruning. Partial key whitening and query-energy ordering construct a representation that balances key scales and concentrates query energy while preserving
attention scores before compression. Experiments show improved accuracy under INT2 quantization and channel pruning, together with higher decoding throughput in the evaluated decoding configurations. The results demonstrate the value of designing query and key coordinates
for quantization and channel selection together.

\newpage

\bibliography{iclr2027_conference}
\bibliographystyle{iclr2027_conference}

\newpage
\appendix

\setlength{\emergencystretch}{2em}

\section{Additional Related Work}
\label{app:additional_related}
\label{app:related}

\noindent\textbf{Vector quantization.}
CommVQ~\citep{commvq} uses additive vector quantization with
codebooks designed to commute with RoPE, allowing reconstruction
to be integrated into attention. NOVA-KV~\citep{novakv} derives
non-orthogonal key transforms under an attention-weighted distortion
criterion. Its transform maps query-weighted key error to Euclidean
error, and it applies vector quantization to groups of transformed
coordinates. Both NOVA-KV and Dual-QK use query statistics to guide
key compression. Dual-QK designs the coordinates for scalar INT2
quantization and dynamic channel selection, with query energy
concentrated in a small subset of channels.

\noindent\textbf{Quantization and low-rank compression.}
GEAR~\citep{gear} combines low-bit quantization with low-rank
correction of quantization residuals and sparse storage of outliers.
STAR-KV~\citep{starkv} learns ranks at attention-head and block
levels through differentiable soft thresholding, and combines
low-rank representations with mixed-precision quantization.
Dual-QK retains the full transformed channel dimension in storage.
Its channel mask controls which key coordinates are read for the
current query, so unselected coordinates remain available at later
steps.

\noindent\textbf{Channel pruning.}
LeanK~\citep{leank} learns static key-channel masks in two stages,
first estimating channel importance and then fitting masks under
sparsity and hardware-alignment constraints. SparK~\citep{spark}
uses query-aware unstructured channel pruning and approximates
omitted key entries during attention using statistics collected
at prefill. These approaches reduce the stored key representation.
Dual-QK instead stores every transformed key channel at low precision
and computes a shared read mask from the current GQA queries for
each bucket.

\noindent\textbf{Learned token eviction.}
KVzap~\citep{kvzap} trains lightweight predictors on hidden states
to estimate per-head token importance. It uses score thresholds
and a protected recent window to prune KV entries during prefill
and decoding. This adapts the stored token budget to the input.
Dual-QK keeps all token positions and reduces the precision and
number of key channels read at each step. Combining it with token
eviction would additionally require deciding which positions to
retain.

\section{Experimental Details}
\label{app:experimental_details}

\subsection{Models and Evaluation}
\label{app:sampling}
\label{app:datasets}

Tables~\ref{tab:app_models} and~\ref{tab:app_datasets} summarize the
models, generation settings, and evaluation datasets. Each generative
run samples one response per problem. Main generative results report
mean $\pm$ standard deviation over five runs for Llama and Qwen3,
and three for Ministral-3.

\begin{table}[b]
\caption{Models and generation settings. Max.\ tokens is the generation
limit per problem. A dash indicates no top-$p$ or top-$k$ filtering.}
\label{tab:app_models}
\centering
\begingroup
\scriptsize
\setlength{\tabcolsep}{2pt}
\renewcommand{\arraystretch}{0.9}
\begin{adjustbox}{max width=\linewidth}
\begin{tabular}{l|ccccc}
\toprule
\textbf{Model} & $\boldsymbol{\alpha}$ & \textbf{Temp.}
& \textbf{Top-$p$} & \textbf{Top-$k$} & \textbf{Max.\ tokens} \\
\midrule
Llama-3.1-8B-Instruct & 0.75 & 0.6 & 0.9 & -- & 16,384 \\
Qwen3-4B-Thinking-2507 & 0.5 & 0.6 & 0.95 & 20 & 32,768 \\
Qwen3-8B & 0.5 & 0.6 & 0.95 & 20 & 32,768 \\
Ministral-3-14B-Reasoning-2512 & 0.375 & 1.0 & -- & -- & 65,536 \\
\bottomrule
\end{tabular}
\end{adjustbox}
\endgroup
\end{table}

\begin{table}[b]
\caption{Evaluation datasets. LiveCodeBench v6 uses the February$\sim$May
2025 subset. LongBench includes eight tasks.}
\label{tab:app_datasets}
\centering
\begingroup
\scriptsize
\setlength{\tabcolsep}{2pt}
\renewcommand{\arraystretch}{0.9}
\begin{adjustbox}{max width=\linewidth}
\begin{tabular}{l|rl}
\toprule
\textbf{Benchmark} & \textbf{Examples} & \textbf{Metric} \\
\midrule
GPQA-Diamond & 198 & Accuracy \\
HumanEval & 164 & Pass@1 \\
LiveCodeBench v6 & 131 & Pass@1 \\
MATH-500 & 500 & Accuracy \\
AIME25 & 30 & Accuracy \\
RULER NIAH & 200 per length & Substring match \\
LongBench & 200$\sim$500 per task & Task-specific \\
\bottomrule
\end{tabular}
\end{adjustbox}
\endgroup
\end{table}

We evaluate eight RULER NIAH subtasks~\citep{ruler} through
lm-evaluation-harness, with 25 examples each at context lengths
from 4K to 128K. We use greedy decoding with a 128-token
limit and average scores equally across subtasks. Results report
mean $\pm$ standard deviation over three independently generated
datasets. LongBench uses greedy decoding and one run per
configuration, following Appendix~\ref{app:longbench_protocol}.

\subsection{Calibration}
\label{app:calibration}

\noindent\textbf{Data and statistics.}
We use the first 128 non-overlapping 2,048-token windows from
WikiText-2 training data~\citep{wikitext2}, totaling 262,144 tokens.
Data are tokenized separately for each model, with the two Qwen3
models sharing a token stream. BF16 forward passes collect post-RoPE
queries and keys, after QK normalization for Qwen3. We accumulate
second moments in FP64 per layer and KV head, pooling query heads
within each GQA group. Transforms are computed in FP32 with
$\epsilon=10^{-5}\operatorname{tr}(C_K)/d$.

\noindent\textbf{Diagonal refinement.}
We fit $D$ per layer and KV head to minimize the query-weighted key
error in Equation~\ref{eq:app_weighted_error}, keeping $\delta_0=1$.
In base coordinates, we sample 2,048 fitting keys and 2,048 validation
keys from a Gaussian with the calibrated key mean and covariance.
We first optimize the logarithm of the uniform-noise approximation
for 400 Adam steps at learning rate 0.06. We then optimize sampled
INT2 error for 300 steps at learning rate 0.02, using a
straight-through estimator for rounding. The free scales are
normalized to unit geometric mean. We retain $D$ only if it reduces
validation error relative to $D=I$, then fold the final scales into
$A$ and $B$.

\noindent\textbf{Whitening strength.}
We select $\alpha$ from $\{0,1/8,\ldots,1\}$ once per model, fitting
$D$ for each candidate. Selection uses 64 inputs not used for
calibration. At each of 8K and 32K tokens, these comprise 16
WikiText-2 windows and 16 synthetic documents containing four
passkeys. We minimize relative squared error of
the final decoder hidden states against BF16, averaged over inputs
and equally over 0\% and 40\% sparsity. Table~\ref{tab:app_alpha_select}
reports the criterion. The selected transforms are reused across
benchmarks and context lengths.

\begin{table}[t]
\caption{Relative hidden-state error (\%) for selecting $\alpha$,
averaged equally over 0\% and 40\% sparsity. Lower is better.
Bold marks the selected value for each model.}
\label{tab:app_alpha_select}
\centering
\begingroup
\scriptsize
\setlength{\tabcolsep}{2pt}
\renewcommand{\arraystretch}{0.9}
\begin{adjustbox}{max width=\linewidth}
\begin{tabular}{c|cccc}
\toprule
$\boldsymbol{\alpha}$ & \textbf{Llama-3.1-8B}
& \textbf{Qwen3-4B} & \textbf{Qwen3-8B} & \textbf{Ministral-3-14B} \\
\midrule
0     & 12.35 & 23.71 & 46.10 & 4.07 \\
0.125 & 9.80 & 12.76 & 30.70 & 3.84 \\
0.25  & 9.62 & 13.98 & 19.11 & 3.71 \\
0.375 & 9.84 & 11.57 & 11.12 & \textbf{3.64} \\
0.5   & 10.44 & \textbf{7.68} & \textbf{8.66} & 3.89 \\
0.625 & 9.32 & 10.80 & 13.46 & 4.10 \\
0.75  & \textbf{9.27} & 15.22 & 16.42 & 4.41 \\
0.875 & 10.13 & 15.55 & 27.24 & 4.65 \\
1     & 10.07 & 21.31 & 47.81 & 5.79 \\
\bottomrule
\end{tabular}
\end{adjustbox}
\endgroup
\end{table}

\noindent\textbf{Value quantization.}
We use OSCAR's calibrated value rotation composed with a Hadamard
transform~\citep{oscar}, fitted on the first 32 calibration windows.
Values use per-token asymmetric INT2 with a clipping ratio of 0.92.
Dual-QK keys are not clipped.

\noindent\textbf{Calibration-domain ablation.}
For Table~\ref{tab:calibration_domain}, we refit all calibrated
components with $\alpha=0.5$ and 2,048-token windows. MMLU uses
1,531 validation prompts across all subjects, forming 98 windows.
For GPQA-Diamond, each prompt contains a question and its four
answer choices. The 198 prompts form 25 windows. These reuse the GPQA evaluation questions,
whereas the default calibration uses WikiText-2. The ablation and
its BF16 reference use three runs on one RTX PRO 6000 Blackwell
Server Edition GPU.

\section{Additional Analysis}
\label{app:analysis}

We analyze the channel statistics and quantization errors in
Figure~\ref{fig:app_channel_profiles}. Moment statistics use the
calibration data. Channel-error measurements use four 2,048-token
WikiText-2 test windows and evaluate each layer separately, without
BF16 sink or recent tokens and without propagating errors across
layers.

\subsection{Dual Transforms}
\label{app:dual_geometry}

For the base pair in Equation~\ref{eq:dual_pair}, the transformed
second moments satisfy
\begin{equation}
 A_0^{\mathsf T}C_QA_0=\Lambda,
 \qquad
 B_0^{\mathsf T}C_KB_0
 =U^{\mathsf T}\bigl(G^{1-\alpha}-\epsilon G^{-\alpha}\bigr)U.
 \label{eq:app_moments}
\end{equation}
The first identity follows from the eigendecomposition defining $U$.
The second uses $C_K=G-\epsilon I$. A key-moment eigenvalue $\mu_i$
therefore becomes $\mu_i(\mu_i+\epsilon)^{-\alpha}$, which approaches
$\mu_i^{1-\alpha}$ when $\mu_i\gg\epsilon$. Partial whitening reduces
the spread of key energies, while $U$ orders the compensated query
energy. These identities use uncentered moments and do not require
zero-mean queries.

\begin{figure}[t]
\centering
\includegraphics[width=\linewidth]{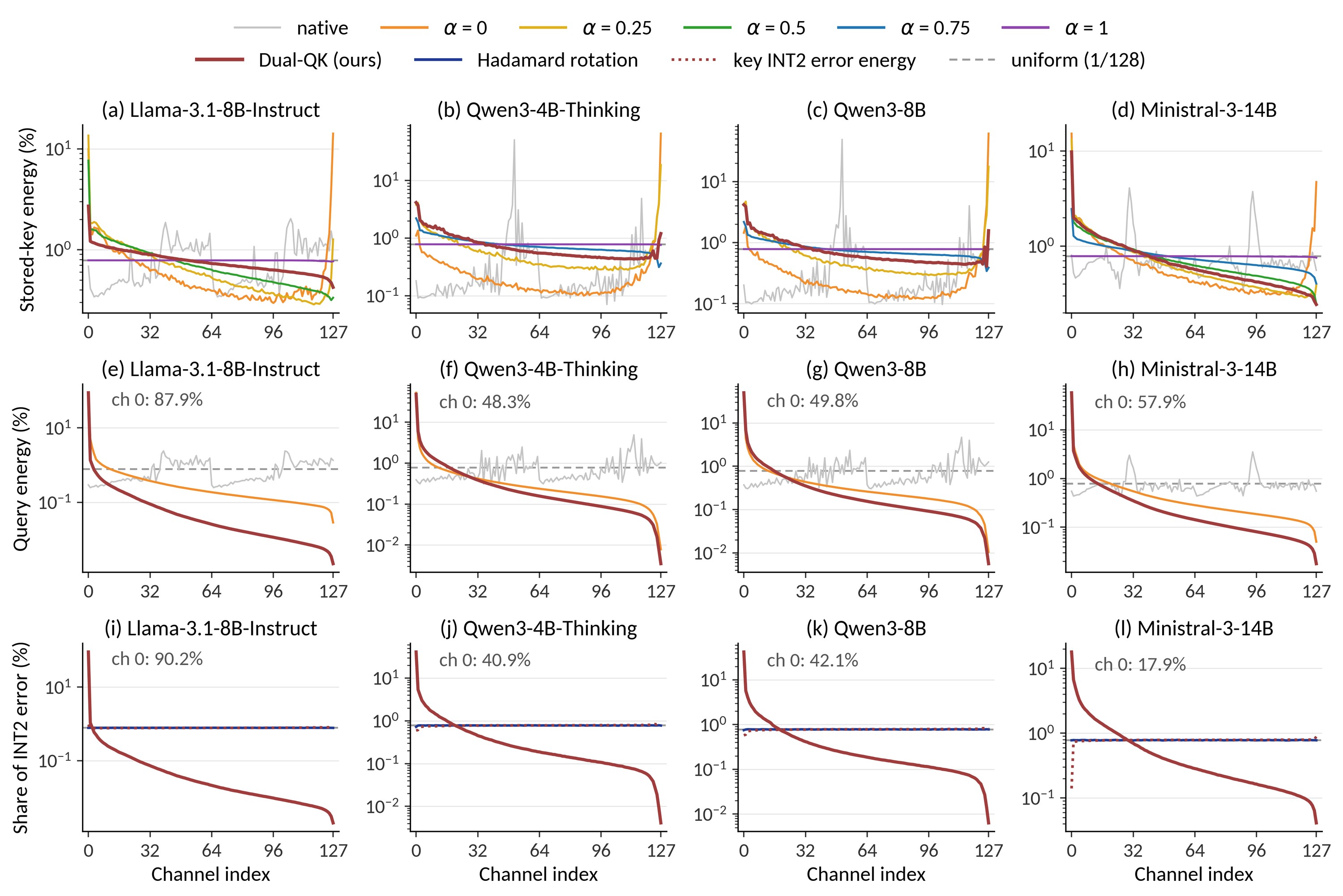}
\caption{Channel profiles across four models. (a$\sim$d) Base key energy
at different $\alpha$. (e$\sim$h) Query energy in the original, Q-PCA,
and deployed dual coordinates. (i$\sim$l) Channel-wise shares of key
reconstruction error (dotted) and query-weighted error terms
(solid), before channel-0 protection, including a Hadamard control.
The first two rows use calibration moments and the last uses
WikiText-2 test activations. Error shares are pooled across layers
and KV heads.}
\label{fig:app_channel_profiles}
\end{figure}

Figure~\ref{fig:app_channel_profiles}(a$\sim$h) shows the key and query
profiles across models. Stronger whitening makes the key energies
more uniform but also changes query scaling. We select $\alpha$
using the hidden-state error in Appendix~\ref{app:calibration} to
account for the combined effect of quantization and pruning.

\subsection{Channel Selection}
\label{app:channel_selection}

The group-L2 rule retains the most query energy for a fixed channel
budget. When the centered key covariance in the selection basis is
$\sigma^2 I$, the variance of the omitted dot-product contribution
for a fixed query is
\begin{equation}
 \operatorname{Var}_{k}\!\left[
   \sum_{j\notin\mathcal M}q_jk_j
 \right]
 =\sigma^2\sum_{j\notin\mathcal M}q_j^2,
 \label{eq:app_pruning_variance}
\end{equation}
where $\mathcal M$ is the retained channel set. Summing over the
query heads in a GQA group gives the group-L2 criterion. Partial whitening of uncentered moments does
not guarantee this covariance condition, so query energy serves as
a proxy for pruning error in Dual-QK.

Selection uses $qT_bA_0$, before diagonal refinement. Scaling query
channel $j$ by $\delta_j$ and its key channel by $1/\delta_j$ leaves
their product unchanged. Computing the mask in base coordinates
therefore makes selection independent of the scaling used to refine
quantization.

\subsection{Key Quantization Error}
\label{app:int2_error}

Let $\xi$ be key quantization error in the stored coordinates.
Under the calibration model in which $\xi$ is independent of the
query, $A^{\mathsf T}C_QA=D\Lambda D$ gives
\begin{equation}
 \mathbb E\bigl[(qA\xi^{\mathsf T})^2\bigr]
 =\sum_j\lambda_j\delta_j^2\,\mathbb E[\xi_j^2].
 \label{eq:app_weighted_error}
\end{equation}
Squared attention-logit error includes an additional factor of
$1/d$. This expression motivates the calibration objective.
It does not model error propagation during autoregressive decoding.

Figure~\ref{fig:app_channel_profiles}(i$\sim$l) compares reconstruction
error with its query-weighted channel contributions. A channel can
have modest reconstruction error but a large contribution to score
error when its query energy is high. Retaining key channel-0 in
BF16 reduces this contribution. Excluding it from range estimation
also narrows the INT2 range when it determines a row extremum.
Diagonal refinement adjusts the remaining scales to reduce the same
query-weighted error. The effect on downstream accuracy is evaluated
in Table~\ref{tab:component_ablation}.

\subsection{Bucket-Relative RoPE}
\label{app:bucket_rope}

For a key at position $n$ in bucket $b$, removing the anchor rotation
leaves position $n-a_b$. Applying the same adjustment to the current
query preserves their relative position and uncompressed score
(Equation~\ref{eq:bucket_score_preservation}). The adjustment
precedes the dual transform because a general channel transform
does not commute with RoPE.

With $w=2048$, stored-key offsets remain within the calibration
window. This bounds their positional offsets but does not guarantee
matching activation statistics. Query offsets relative to older
buckets can still exceed the window. Anchor adjustments use each
checkpoint's rotary frequencies, including YaRN for Qwen3-8B beyond
32K and for Ministral. Table~\ref{tab:component_ablation} evaluates
the effect of removing bucket-relative RoPE.

\section{Additional Results}
\label{app:more_results}

\subsection{Results without Channel Pruning}
\label{app:results_s0}

\begin{table}[t]
\caption{Generative accuracy (\%) without query-channel pruning.
Scores report mean $\pm$ standard deviation over five seeds.
Capacity and bandwidth are effective bits per KV element at
128K context. Bold marks the highest mean among quantized methods
within each model.}
\label{tab:app_s0}
\footnotesize
\centering
\setlength{\tabcolsep}{1.5pt}
\renewcommand{\arraystretch}{0.9}
\resizebox{\linewidth}{!}{%
\begin{tabular}{cl|cc|ccccc}
\toprule
\multirow{2}{*}{\textbf{~~Model~~}}
& \multirow{2}{*}{\textbf{Method}}
& \multicolumn{2}{c|}{\textbf{KV bits per element}}
& \multicolumn{5}{c}{\textbf{Benchmark score (\%)}} \\
\cmidrule{3-9}
& & \textbf{Capacity} & \textbf{Bandwidth}
& \textbf{GPQA-Diamond} & \textbf{HumanEval}
& \textbf{LCB v6} & \textbf{MATH-500} & \textbf{AIME25} \\
\midrule

\multirow{5}{*}{%
    \fontsize{8}{8}\selectfont
    \shortstack{\textbf{Llama-3.1-8B}\\\textbf{-Instruct}}%
}
& BF16 & 16 & 16
& \dqscore{26.2}{1.5}
& \dqscore{66.7}{2.5}
& \dqscore{16.5}{1.4}
& \dqscore{49.4}{1.5}
& \dqscore{1.3}{1.8} \\
& TurboQuant & 3.13 & 3.13
& \dqscorebest{26.8}{3.1}
& \dqscore{59.3}{2.9}
& \dqscore{12.5}{1.3}
& \dqscore{39.9}{1.1}
& \dqscore{0.0}{0.0} \\
& KIVI-2 & 3.01 & 3.01
& \dqscore{23.8}{2.8}
& \dqscore{61.7}{1.7}
& \dqscore{14.8}{1.8}
& \dqscore{43.6}{1.0}
& \dqscore{0.0}{0.0} \\
& OSCAR & 2.28 & 2.28
& \dqscore{24.3}{1.7}
& \dqscore{65.6}{3.0}
& \dqscore{15.3}{0.9}
& \dqscore{50.8}{0.5}
& \dqscorebest{2.0}{1.8} \\
& \textbf{Dual-QK (ours)} & 2.34 & 2.34
& \dqscore{26.4}{1.9}
& \dqscorebest{66.0}{1.7}
& \dqscorebest{16.5}{0.7}
& \dqscorebest{50.9}{0.6}
& \dqscore{1.3}{3.0} \\
\midrule

\multirow{5}{*}{%
    \fontsize{8}{8}\selectfont
    \shortstack{\textbf{Qwen3-4B}\\\textbf{-Thinking-2507}}%
}
& BF16 & 16 & 16
& \dqscore{65.6}{1.9}
& \dqscore{96.2}{0.7}
& \dqscore{54.0}{1.6}
& \dqscore{98.0}{0.3}
& \dqscore{74.7}{3.8} \\
& TurboQuant & 3.13 & 3.13
& \dqscore{28.7}{1.9}
& \dqscore{15.9}{2.0}
& \dqscore{0.2}{0.3}
& \dqscore{36.2}{2.0}
& \dqscore{1.3}{1.8} \\
& KIVI-2 & 3.01 & 3.01
& \dqscore{62.5}{1.3}
& \dqscore{93.4}{1.2}
& \dqscorebest{47.8}{1.8}
& \dqscorebest{96.8}{0.4}
& \dqscorebest{64.7}{5.1} \\
& OSCAR & 2.28 & 2.28
& \dqscore{63.5}{1.6}
& \dqscorebest{96.1}{1.0}
& \dqscore{46.7}{1.3}
& \dqscore{96.4}{0.5}
& \dqscorebest{64.7}{3.8} \\
& \textbf{Dual-QK (ours)} & 2.34 & 2.34
& \dqscorebest{65.8}{1.4}
& \dqscore{95.9}{1.1}
& \dqscore{42.9}{1.0}
& \dqscore{95.8}{0.3}
& \dqscore{54.0}{4.9} \\
\midrule

\multirow{5}{*}{%
    \fontsize{8}{8}\selectfont
    \shortstack{\textbf{Qwen3-8B}}%
}
& BF16 & 16 & 16
& \dqscore{56.1}{2.2}
& \dqscore{90.1}{1.3}
& \dqscore{50.2}{2.9}
& \dqscore{96.4}{0.3}
& \dqscore{65.3}{3.0} \\
& TurboQuant & 3.13 & 3.13
& \dqscore{47.1}{2.2}
& \dqscore{75.7}{1.6}
& \dqscore{24.7}{1.7}
& \dqscore{92.4}{0.6}
& \dqscore{50.7}{2.8} \\
& KIVI-2 & 3.01 & 3.01
& \dqscore{54.2}{1.9}
& \dqscore{91.8}{0.9}
& \dqscorebest{46.6}{1.4}
& \dqscorebest{95.9}{0.3}
& \dqscorebest{64.0}{6.4} \\
& OSCAR & 2.28 & 2.28
& \dqscorebest{56.6}{1.2}
& \dqscorebest{92.2}{1.1}
& \dqscore{46.1}{1.6}
& \dqscore{95.8}{0.5}
& \dqscore{61.3}{1.8} \\
& \textbf{Dual-QK (ours)} & 2.34 & 2.34
& \dqscore{55.9}{2.2}
& \dqscore{92.0}{0.9}
& \dqscore{43.2}{2.4}
& \dqscore{95.2}{0.4}
& \dqscore{61.3}{3.0} \\
\bottomrule
\end{tabular}%
}
\end{table}

Table~\ref{tab:app_s0} reports the full results at 0\% query-channel
sparsity, extending Table~\ref{tab:zero_sparsity} with KIVI-2 and
Qwen3-4B-Thinking-2507.
Dual-QK maintains competitive accuracy without pruning, with
five-benchmark averages within one percentage point of OSCAR on
Llama-3.1-8B-Instruct and Qwen3-8B.
On Qwen3-4B-Thinking-2507, Dual-QK also remains close to BF16 on
GPQA-Diamond and HumanEval, although larger gaps remain on
LiveCodeBench and AIME25.

\subsection{LongBench Results}
\label{app:longbench_protocol}
\label{app:results_longbench}

We evaluate the full test sets of eight LongBench~\citep{longbench}
tasks using greedy decoding and task-specific generation limits.
Inputs exceeding 31.5K tokens are truncated in the middle.
We apply chat templates to Qasper, QMSum, and MultiNews, and disable
thinking for Qwen3-8B. For TREC, prompts end with the answer prefix
and the first output line is scored following the official evaluator.

Tables~\ref{tab:app_longbench} and~\ref{tab:app_longbench_s0}
compare accuracy with and without query-channel pruning. At 40\% sparsity, Dual-QK achieves mean scores
of 37.68 and 39.41 on Llama-3.1-8B-Instruct and Qwen3-8B,
exceeding OSCAR by 10.08 and 3.12 points, respectively.
Increasing sparsity from 0\% to 40\% reduces the Dual-QK mean by
only 1.54 and 1.22 points, compared with 11.32 and 4.09 for OSCAR.
Without pruning, Dual-QK also achieves the highest mean among the
quantized methods on both models.

\begin{table*}[t]
\caption{LongBench scores with 40\% query-channel sparsity for
quantized methods. BF16 is unpruned. Mean averages the eight task
scores. Bold marks the highest score among quantized methods
within each model.}
\label{tab:app_longbench}
\centering
\begingroup
\fontsize{7}{8}\selectfont
\setlength{\tabcolsep}{2pt}
\renewcommand{\arraystretch}{0.95}
\begin{tabular*}{\textwidth}{@{\extracolsep{\fill}}cl|cccccccc|c@{}}
\toprule
\textbf{Model}
& \textbf{Method}
& \textbf{Qasper}
& \textbf{QMSum}
& \shortstack{\textbf{Multi}\\\textbf{News}}
& \textbf{TREC}
& \shortstack{\textbf{Trivia}\\\textbf{QA}}
& \textbf{SAMSum}
& \textbf{LCC}
& \shortstack{\textbf{Repo}\\\textbf{Bench-P}}
& \textbf{Mean} \\
\midrule

\multirow{4}{*}{%
    \shortstack{\textbf{Llama-3.1-8B}\\\textbf{-Instruct}}%
}
& BF16
& 39.59 & 23.79 & 24.87 & 56.00 & 80.44
& 33.84 & 35.94 & 30.01 & 40.56 \\
& TurboQuant
& 14.30 & 18.66 & 12.99 & 46.50 & 54.04
& 20.86 & 18.36 & 17.99 & 25.46 \\
& OSCAR
& 18.03 & 18.29 & 15.45 & 49.50 & 57.49
& 24.26 & 18.13 & 19.62 & 27.60 \\
& \textbf{Dual-QK (ours)}
& \textbf{38.97} & \textbf{22.99} & \textbf{24.31}
& \textbf{54.00} & \textbf{75.00} & \textbf{33.60}
& \textbf{26.95} & \textbf{25.65} & \textbf{37.68} \\
\midrule

\multirow{4}{*}{\textbf{Qwen3-8B}}
& BF16
& 46.72 & 23.80 & 23.86 & 58.63 & 89.13
& 41.04 & 25.98 & 23.51 & 41.58 \\
& TurboQuant
& 2.50 & 6.72 & 11.74 & 23.50 & 0.04
& 1.23 & 14.48 & 12.69 & 9.11 \\
& OSCAR
& 34.15 & 20.28 & 22.76 & \textbf{65.50} & 81.36
& 36.11 & 15.03 & 15.15 & 36.29 \\
& \textbf{Dual-QK (ours)}
& \textbf{40.78} & \textbf{22.34} & \textbf{23.56}
& 61.00 & \textbf{87.07} & \textbf{38.45}
& \textbf{21.76} & \textbf{20.27} & \textbf{39.41} \\
\bottomrule
\end{tabular*}
\endgroup
\end{table*}

\begin{table*}[t]
\caption{LongBench scores without query-channel pruning.
Mean averages the eight task scores. Bold marks the highest
score among quantized methods within each model.}
\label{tab:app_longbench_s0}
\centering
\begingroup
\fontsize{7}{8}\selectfont
\setlength{\tabcolsep}{2pt}
\renewcommand{\arraystretch}{0.95}
\begin{tabular*}{\textwidth}{@{\extracolsep{\fill}}cl|cccccccc|c@{}}
\toprule
\textbf{Model}
& \textbf{Method}
& \textbf{Qasper}
& \textbf{QMSum}
& \shortstack{\textbf{Multi}\\\textbf{News}}
& \textbf{TREC}
& \shortstack{\textbf{Trivia}\\\textbf{QA}}
& \textbf{SAMSum}
& \textbf{LCC}
& \shortstack{\textbf{Repo}\\\textbf{Bench-P}}
& \textbf{Mean} \\
\midrule

\multirow{4}{*}{%
    \shortstack{\textbf{Llama-3.1-8B}\\\textbf{-Instruct}}%
}
& BF16
& 39.59 & 23.79 & 24.87 & 56.00 & 80.44
& 33.84 & 35.94 & 30.01 & 40.56 \\
& TurboQuant
& \textbf{39.21} & 22.60 & 22.94 & 55.25 & 75.95
& 30.26 & 27.67 & 25.68 & 37.44 \\
& OSCAR
& 36.91 & \textbf{22.67} & 23.99 & 53.50 & 76.53
& \textbf{34.10} & \textbf{35.33} & 28.31 & 38.92 \\
& \textbf{Dual-QK (ours)}
& 36.06 & 22.65 & \textbf{24.67} & \textbf{56.50}
& \textbf{77.32} & 33.87 & 34.28
& \textbf{28.38} & \textbf{39.22} \\
\midrule

\multirow{4}{*}{\textbf{Qwen3-8B}}
& BF16
& 46.72 & 23.80 & 23.86 & 58.63 & 89.13
& 41.04 & 25.98 & 23.51 & 41.58 \\
& TurboQuant
& 37.08 & 20.96 & \textbf{23.99} & \textbf{66.50}
& 73.78 & 36.17 & 20.99 & 19.88 & 37.42 \\
& OSCAR
& \textbf{43.73} & 23.07 & 23.53 & 55.43
& \textbf{89.78} & 40.23 & 24.80 & \textbf{22.51} & 40.38 \\
& \textbf{Dual-QK (ours)}
& 41.34 & \textbf{23.23} & 23.71 & 61.70 & 88.77
& \textbf{40.62} & \textbf{24.89} & 20.79 & \textbf{40.63} \\
\bottomrule
\end{tabular*}
\endgroup
\end{table*}

\subsection{Comparison with ThinK}
\label{app:results_think}

We compare Dual-QK with KIVI-2 and OSCAR combined with
ThinK~\citep{think} at 40\% channel sparsity.
ThinK removes key channels using a prompt-derived mask while keeping
recent tokens unpruned. Dual-QK retains all key channels and selects
the channels to read for each query. We report both capacity and
bandwidth costs to account for this difference.

On LongBench (Table~\ref{tab:app_think_longbench}), Dual-QK exceeds
OSCAR + ThinK by 5.56 and 6.58 mean-score points on the two models.
Its means are within 0.53 points of KIVI-2 + ThinK, with lower
bandwidth cost at 1.93 versus 2.41 bits per KV element.
On the generative benchmarks (Table~\ref{tab:app_think}), Dual-QK
achieves higher mean GPQA-Diamond and HumanEval accuracy than both
ThinK combinations on both models.

\begin{table*}[t]
\caption{LongBench scores with ThinK and Dual-QK at 40\% channel
sparsity. BF16 is unpruned. Capacity and bandwidth are effective
bits per KV element at 128K context. Mean averages the eight task
scores. Bold marks the highest score among compressed methods
within each model.}
\label{tab:app_think_longbench}
\centering
\scriptsize
\setlength{\tabcolsep}{1.5pt}
\renewcommand{\arraystretch}{0.9}
\begin{adjustbox}{max width=\textwidth}
\begin{tabular}{cl|cc|cccccccc|c}
\toprule
\multirow{2}{*}{\textbf{Model}}
& \multirow{2}{*}{\textbf{Method}}
& \multicolumn{2}{c|}{\textbf{KV bits per element}}
& \multicolumn{9}{c}{\textbf{LongBench score}} \\
\cmidrule{3-13}
& & \textbf{Capacity} & \textbf{Bandwidth} & \textbf{Qasper} & \textbf{QMSum} & \shortstack{\textbf{Multi}\\\textbf{News}} & \textbf{TREC} & \shortstack{\textbf{Trivia}\\\textbf{QA}} & \textbf{SAMSum} & \textbf{LCC} & \shortstack{\textbf{Repo}\\\textbf{Bench-P}} & \textbf{Mean} \\
\midrule
\multirow{4}{*}{\shortstack{\textbf{Llama-3.1-8B}\\\textbf{-Instruct}}} & BF16 & 16 & 16 & 39.59 & 23.79 & 24.87 & 56.00 & 80.44 & 33.84 & 35.94 & 30.01 & 40.56 \\
 & KIVI-2 + ThinK & 2.41 & 2.41 & 38.00 & \textbf{23.58} & 21.67 & \textbf{54.54} & \textbf{78.83} & 30.59 & 31.98 & 26.48 & \textbf{38.21} \\
 & OSCAR + ThinK & 1.89 & 1.89 & 19.74 & 20.98 & 17.49 & 33.00 & 70.12 & \textbf{34.68} & \textbf{32.64} & \textbf{28.32} & 32.12 \\
 & \textbf{Dual-QK (ours)} & 2.34 & 1.93 & \textbf{38.97} & 22.99 & \textbf{24.31} & 54.00 & 75.00 & 33.60 & 26.95 & 25.65 & 37.68 \\
\midrule
\multirow{4}{*}{\shortstack{\textbf{Qwen3-8B}}} & BF16 & 16 & 16 & 46.72 & 23.80 & 23.86 & 58.63 & 89.13 & 41.04 & 25.98 & 23.51 & 41.58 \\
 & KIVI-2 + ThinK & 2.41 & 2.41 & \textbf{42.71} & \textbf{23.24} & 23.54 & 54.82 & \textbf{87.18} & \textbf{40.05} & 24.43 & \textbf{21.99} & \textbf{39.74} \\
 & OSCAR + ThinK & 1.89 & 1.89 & 26.82 & 20.33 & 16.46 & 30.75 & 84.87 & 37.25 & \textbf{24.46} & 21.72 & 32.83 \\
 & \textbf{Dual-QK (ours)} & 2.34 & 1.93 & 40.78 & 22.34 & \textbf{23.56} & \textbf{61.00} & 87.07 & 38.45 & 21.76 & 20.27 & 39.41 \\
\bottomrule
\end{tabular}
\end{adjustbox}
\end{table*}

\begin{table}[t]
\caption{Generative accuracy (\%) with ThinK and Dual-QK at 40\%
channel sparsity. BF16 is unpruned. Scores report mean $\pm$
standard deviation over three seeds. Bold marks the highest mean
among compressed methods within each model.}
\label{tab:app_think}
\centering
\scriptsize
\setlength{\tabcolsep}{4pt}
\renewcommand{\arraystretch}{0.9}
\begin{adjustbox}{max width=\linewidth}
\begin{tabular}{cl|cc}
\toprule
\multirow{2}{*}{\textbf{Model}}
& \multirow{2}{*}{\textbf{Method}}
& \multicolumn{2}{c}{\textbf{Benchmark score (\%)}} \\
\cmidrule{3-4}
& & \textbf{GPQA-Diamond} & \textbf{HumanEval} \\
\midrule
\multirow{4}{*}{\shortstack{\textbf{Llama-3.1-8B}\\\textbf{-Instruct}}} & BF16 & \dqscore{26.8}{1.5} & \dqscore{67.7}{2.4} \\
 & KIVI-2 + ThinK & \dqscore{22.2}{2.3} & \dqscore{57.7}{1.8} \\
 & OSCAR + ThinK & \dqscore{14.5}{1.5} & \dqscore{62.8}{0.6} \\
 & \textbf{Dual-QK (ours)} & \dqscorebest{24.2}{1.5} & \dqscorebest{65.2}{3.4} \\
\midrule
\multirow{4}{*}{\shortstack{\textbf{Qwen3-8B}}} & BF16 & \dqscore{56.9}{1.9} & \dqscore{89.2}{0.3} \\
 & KIVI-2 + ThinK & \dqscore{55.7}{2.0} & \dqscore{88.2}{0.4} \\
 & OSCAR + ThinK & \dqscore{2.0}{0.9} & \dqscore{35.6}{0.4} \\
 & \textbf{Dual-QK (ours)} & \dqscorebest{56.7}{1.8} & \dqscorebest{91.9}{0.9} \\
\bottomrule
\end{tabular}
\end{adjustbox}
\end{table}

\end{document}